%% file: main.tex
\documentclass[conference]{IEEEtran}
\IEEEoverridecommandlockouts

\usepackage{listings}
\usepackage{hyperref}
\usepackage{graphicx}
\usepackage{amsmath,amssymb,amsfonts}
\usepackage{algorithmic}
\usepackage{textcomp}
\usepackage{xcolor}
\usepackage{acronym}
\usepackage{todonotes}
\usepackage{booktabs}
\usepackage{adjustbox}
\usepackage{csvsimple}
\usepackage{subcaption}

\usepackage[utf8]{inputenc} 
\usepackage[T1]{fontenc}    
\usepackage{url}            
\usepackage{nicefrac}       
\usepackage{cleveref}

\acrodef{RDF}[RDF]{Resource Description Framework}
\acrodef{LLM}[LLM]{Large Language Model}
\acrodef{DP}[DP]{Differential Privacy}
\acrodef{FL}[FL]{Federated learning}
\acrodef{ML}[ML]{Machine Learning}
\acrodef{PHT}[PHT]{Personal Health Train}
\acrodef{PATE}[PATE]{Private Aggregation of Teacher Ensemble}
\acrodef{JS}[JS]{Jaccard Similarity}

\newcommand{\NAME}{{PrivFusion}}

\usepackage{cite}
\def\BibTeX{{\rm B\kern-.05em{\sc i\kern-.025em b}\kern-.08em
    T\kern-.1667em\lower.7ex\hbox{E}\kern-.125emX}}
\begin{document}

\title{PrivFusion: A Privacy-preserving Multi-Agent Framework for Harmonizing Distributed Datasets}

\author{\IEEEauthorblockN{Anisa Halimi}
\IEEEauthorblockA{
\textit{IBM Research}\\
Dublin, Ireland \\
anisa.halimi@ibm.com}
\and
\IEEEauthorblockN{Liubov Nedoshivina}
\IEEEauthorblockA{
\textit{IBM Research}\\
Dublin, Ireland \\
liubov.nedoshivina@ibm.com}
\and
\IEEEauthorblockN{Kieran Fraser}
\IEEEauthorblockA{
\textit{IBM Research}\\
Dublin, Ireland \\
kieran.fraser@ibm.com}
\and
\IEEEauthorblockN{Stefano Braghin}
\IEEEauthorblockA{
\textit{IBM Research}\\
Dublin, Ireland \\
stefanob@ie.ibm.com}
}

\maketitle

\begin{abstract}
The growing availability of clinical data has increased the use of machine learning, yet centralized data aggregation is often infeasible for sensitive health information. Federated Learning (FL) offers a distributed alternative, but its adoption is limited by substantial heterogeneity across institutional datasets, making harmonization a critical but frequently overlooked prerequisite for multi-site analytics. We introduce PrivFusion, a privacy-preserving multi-agent framework that automates the harmonization of structured datasets prior to federated training. PrivFusion uses agents to analyze local data, cluster semantically similar features across sites, and provide iterative transformation recommendations until alignment is achieved. Evaluation across four heterogeneous COVID-19 datasets demonstrates that PrivFusion effectively and efficiently harmonizes multi-site data while substantially reducing manual effort.
\end{abstract}


\input{sections/introduction}
\input{sections/method}
\input{sections/experimental_evaluation}
\input{sections/discussion}
\input{sections/related_work}
\input{sections/conclusions}

\section*{Acknowledgments}\label{sec:acknowledgments}
Anisa Halimi and Stefano Braghin were partly supported by the Innovative Health Initiative Joint Undertaking (IHI JU) under grant agreement No. 101172997 – SEARCH. The content is solely the responsibility of the authors and does not necessarily represent the official views of the agencies funding the research.

\bibliographystyle{ieeetr}
\bibliography{bibliography}

\end{document}

%% file: sections/introduction.tex
\section{Introduction}\label{sec:intro}
\vspace{-3pt}

The volume and complexity of data generated across clinical, biomedical, and public health settings continue to grow rapidly. This has led to the adoption of \ac{ML} methods for tasks such as risk prediction, phenotyping, and cohort discovery~\cite{chen2017disease}. However, traditional \ac{ML} pipelines typically rely on centralized data aggregation, which is often infeasible for sensitive information such as electronic health records or financial data due to privacy, security, and governance constraints. As a result, many healthcare institutions remain unable or unwilling to share raw data, limiting opportunities for large-scale, multi-site analytics. 

\ac{FL} has emerged as an alternative that enables collaborative model training across institutions while keeping data on-premises~\cite{antunes2022federated}. Although \ac{FL} has demonstrated strong potential in various areas, deploying \ac{FL} in the real world remains challenging. A central barrier is the data heterogeneity across institutions. Clinical datasets often differ in feature definitions, coding systems, granularity, measurement practices, and data quality. Before any distributed analysis can be performed, these datasets must be harmonized -- a labor-intensive process that requires aligning variables, resolving semantic inconsistencies, and standardizing representations~\cite{kumar2021data}. Yet, most \ac{FL} methodologies implicitly assume that harmonization has been completed prior to model training, an assumption that is not realistic. As a result, this has become a major bottleneck, delaying or limiting the scalability of federated studies.

To mitigate these challenges, several studies have explored ontology-driven and semantic techniques for data harmonization~\cite{kourou2018cohort,carmen2016generation,chondrogiannis2019novel}. While these approaches alleviate aspects of data heterogeneity, they often require some degree of centralization or rely on predefined common formalisms and mappings, which can be difficult to establish across institutions with differing workflows and data models. These limitations underscore the need for automated, scalable, and privacy-preserving harmonization methods that operate in distributed environments without centralizing sensitive data. Such capabilities would substantially reduce the manual burden on participating institutions, broaden access to collaborative analytics, and ultimately improve the generalizability of resulting models.

To address this gap, we introduce \NAME{}\footnote{Our code is available at https://github.com/IBM/PrivFusion}, a privacy-preserving multi-agent framework designed to harmonize structured datasets across institutions prior to federated model training. In \NAME{}, each participating site, \emph{researcher}, first performs a local analysis of its dataset using a suite of agents that extract data types, generate dataset and feature-level descriptions, infer relationships among features, and generate a small set of synthetic samples. Sites then share this metadata with a central server, \emph{aggregator}. The server uses the received metadata to cluster semantically related features across datasets and, based on these clusters, invokes an additional agent to generate harmonization recommendations for each site. These recommendations specify which features require transformation and the target representations needed for alignment. Each site applies the proposed transformations locally, after which the updated metadata are resent to the server. This iterative process continues until no additional transformations are required. We evaluate \NAME{} on four real-world COVID-19 datasets originating from different countries and find that it typically achieves harmonization within 2–3 iterations across dataset pairs, even when varying the underlying \ac{LLM}. We further observe a consistent increase in feature-name similarity over successive iterations, indicating progressive convergence of the schemas.

%% file: sections/method.tex
\section{System and Threat Models}\label{sec:system_threat_model}
In this section, we introduce our system and threat models.

\noindent\textbf{\textit{System model.}}
We consider a system that includes two parties: (i) two or more researchers, and (ii) a server. The researchers' goal is to collaboratively train a medical diagnostic model with high-quality data. For that, they need to ensure that their data is harmonized in a privacy-preserving way. All computations are outsourced to the server. To harmonize the features across the datasets, each researcher provides some metadata, as will be discussed in Section~\ref{sec:method}, to the server. Using the received metadata, the server determines for each dataset which features require transformations and the target representation, and sends these recommendations to each researcher. The researchers transform their datasets using the received recommendations and conduct their desired study.

\noindent\textbf{\textit{Threat model.}}
We assume honest researchers with legitimate research datasets in an honest-but-curious setting, where the server follows the protocol, but it might leverage shared data and metadata to infer sensitive information about the individuals represented in the datasets. Some of the known attacks are membership inference~\cite{stadler2022synthetic,annamalai2024you} and attribute inference attacks~\cite{ganev2025inadequacy}. In a membership inference attack, the attacker (the server) aims to determine whether a target individual is part of the dataset or not. In an attribute inference attack, the attacker's goal is to infer some additional sensitive information about a target individual given the observed ones.

\section{Proposed Framework}\label{sec:method}
\begin{figure}[tb]
    \centering
    \includegraphics[width=.95\columnwidth]{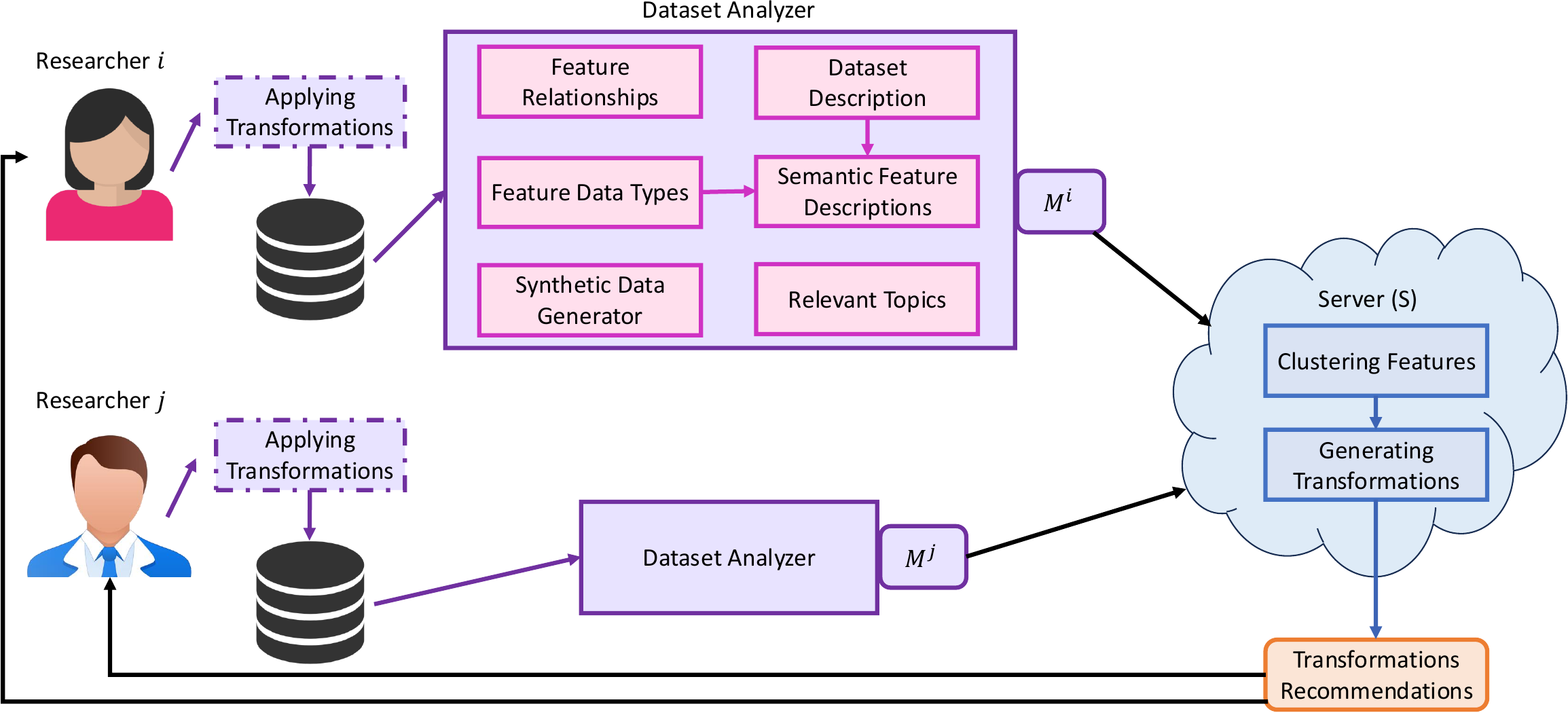}
    \caption{An overview of \NAME{}.}
    \label{fig:architecture}
    \vspace{-10pt}
\end{figure}

To harmonize the features across different datasets, the proposed approach requires researchers to share information about their respective datasets and samples with the server in a privacy-preserving manner. The researchers' goal is to provide metadata useful for determining the required dataset transformations. At the same time, they want to ensure that the metadata does not increase the baseline privacy risk of sharing aggregate statistics about the dataset. Let $S$ denote the server, $D^i$ represent the original dataset of researcher $R^i$, and $M^i$ the metadata that each researcher sends to the server. In the following, we describe the different steps of the proposed framework, shown in \Cref{fig:architecture}.

\noindent\textbf{\textit{Dataset Analyzer.}}
Initially, each researcher locally extracts the feature data types (e.g., string, numeric, floating point single precision, etc.) for their dataset.  Moreover, each feature is associated with a semantic type using advanced data classification techniques~\cite{nedoshivina2024pragmatic,braghin2019extensible} and by mapping against semantic concepts within a selected general purpose ontology. In the following, we leverage DBPedia as our reference ontology. After that, the researchers use an agent $A_D$ to generate a description of the dataset. Given the dataset description and feature data types, each researcher uses another agent, $A_F$, to obtain a semantic description of the features. In addition, each researcher uses an agent $A_T$ to extract a list of relevant topics represented in their dataset $D^i$. To identify feature dependencies, they use an agent $A_R$ that infers relationships between the features. Finally, each researcher generates $n$ synthetic samples, low-utility preserving samples, which will mainly guide the server on harmonizing the features across datasets in the next steps. The researchers' goal is to generate synthetic samples that maintain the format of the features in their respective datasets and whose values are within the domain. Note that, to generate synthetic samples, the researchers can use any of the SOTA \ac{DP} approaches. 

At the end of this stage, each researcher generates a metadata $M^i$ consisting of (i) a description of the dataset, (ii) feature names and data types, (iii) semantic descriptions of individual features, (iv) a list of dataset topics, and (v) inferred relationships among features. In addition to these, each researcher shares $n$ synthetically generated samples. Each researcher, $R^i$, sends their prepared metadata, $M^i$, to the server $S$, which determines which features need to be transformed and their granularity, and then sends this information back to the researchers. 

\noindent\textbf{\textit{Clustering Features.}}
By examining the metadata $M^i$ (specifically, feature names, data types, and semantic types), the server clusters the features across the datasets, grouping those that are semantically similar or related and are likely candidates for merging into a single feature. This is achieved through an agent $A_C$, which clusters features that share similar semantics and are candidates for consolidation. The agent outputs a set of clusters $C$ where each cluster $C_j$ consists of a cluster identifier, feature name, and its dataset name. Features within the same cluster share a cluster ID, indicating potential alignment across datasets. Thus, clusters contain one or more features. 

\noindent\textbf{\textit{Transformations Recommender.}}
Given the clusters obtained in the previous step and the metadata $M^i$ provided by each researcher -- especially, the dataset description, the features semantic type (i.e., DBpedia URIs) and semantic descriptions, the feature relationships, and the synthetic samples -- the server invokes another agent $A_H$ to determine which features within each dataset should be combined, removed, or modified. For each cluster, $A_H$ compares the DBpedia URIs and semantic feature descriptions to converge on a representative DBpedia URI that best reflects the shared concept across datasets, while preserving an appropriate level of granularity. The outcome is a set of transformation recommendations for each researcher, specifying which features require modification and the target representation that the server $S$ sends to the respective parties.

\noindent\textbf{\textit{Applying Transformations.}}
Upon receiving the transformation instructions, each researcher $R^i$ applies the recommended transformations to their dataset. Researchers can implement these transformations directly using Python scripts or leverage a code generation \ac{LLM} $M_C$ to automatically produce the transformation functions. Once the transformations are applied, all participating datasets are harmonized to a common semantic and structural format. This alignment enables collaborative development and training of high-quality diagnostic models across institutions. These steps are repeated until the server does not suggest further transformations or the maximum number of iterations $T$ is reached.

%% file: sections/experimental_evaluation.tex
\section{Evaluation}\label{sec:experiments}
Our evaluation of \NAME{} focuses on two key aspects: (i) the number of iterations required to reach convergence and (ii) the number of feature-level transformations recommended during the harmonization process. We measure feature similarity between datasets at each iteration to showcase harmonization progress and examine how the choice of the underlying \ac{LLM} influences these outcomes.

\noindent\textbf{Datasets.} 
We use four publicly available COVID-19 datasets: (i) COVID-19 Indonesia\footnote{\url{https://www.kaggle.com/dsv/4214699}} (IDN), (ii) COVID-19 Afghanistan\footnote{\url{https://www.kaggle.com/datasets/georgesaavedra/covid19-dataset}}(AFG), (iii) COVID-19 Italy\footnote{\url{https://www.kaggle.com/datasets/sudalairajkumar/covid19-in-italy}}(IT), and (iv) COVID-19 US\footnote{\url{https://www.kaggle.com/dsv/13711559}}(US). Each dataset contains information about COVID-19 cases in the specified country (e.g., the total number of coronavirus cases on a given date in a given location). These datasets share some features in different representations, such as date, ISO code, location, etc. Note that we removed some of the unique features from the AFG and IDN datasets in order to focus the analysis on features that may be aligned.

\textit{Date.} All four datasets include a date feature, but use different formats. IDN records dates as \texttt{mm/dd/yyyy} (\texttt{Date}), while AFG and US use the format \texttt{yyyy-mm-dd} (\texttt{date}). IT provides the most detailed representation, following the ISO-8601 timestamp format \texttt{YYYY-MM-DDTHH:MM:SS} (\texttt{date}).

\textit{Location.} Location granularity and encoding differ widely. US reports \texttt{county} and \texttt{state} level identifiers. IDN uses ISO codes (\texttt{Location ISO Code}) and province-level labels (\texttt{Location}). AFG describes location using 3 features \texttt{iso\_code}, \texttt{continent}, and \texttt{location}. IT offers the finest granularity including \texttt{Country}, \texttt{RegionCode}, \texttt{RegionName}, \texttt{ProvinceName}, \texttt{ProvinceAbbreviation}, \texttt{Lattitude}, \texttt{Longitude}.

\textit{Case counts.} All datasets report the total number of COVID-19 cases, but under different feature names: \texttt{TotalPositiveCases} (Italy), \texttt{cases} (US), \texttt{total\_cases} (Afghanistan), \texttt{Total Cases} (Indonesia).

\begin{table*}[tb]
\centering
\caption{Synthetic samples generated from COVID-19 datasets for Afghanistan, Indonesia, Italy, and US.}
\label{tab:covid-samples}

\scriptsize
\begin{subtable}[t]{0.48\textwidth}
    \centering
    \adjustbox{max width=\textwidth}{%
        \begin{filecontents*}{data/covid_dataset_afg_synthetic_head.csv}
iso_code,continent,location,date,total_cases,population_density
AFG,Asia,Afghanistan,2020-02-24,5.0,54.422
AFG,Asia,Afghanistan,2020-02-25,5.0,54.422
AFG,Asia,Afghanistan,2020-02-26,5.0,54.422
AFG,Asia,Afghanistan,2020-02-27,5.0,54.422
AFG,Asia,Afghanistan,2020-02-28,5.0,54.422
\end{filecontents*}
\csvautobooktabular[separator=comma, respect all, filter={\value{csvrow}<3}]{data/covid_dataset_afg_synthetic_head.csv}%
    }
    \vspace{1em}
    \caption{COVID-19 Afghanistan}
    \label{tab:afg-samples}
\end{subtable}
\hfill
\begin{subtable}[t]{0.48\textwidth}
    \centering
    \adjustbox{max width=\textwidth}{%
        \begin{filecontents*}{data/covid_dataset_ind_synthetic_head.csv}
Date,Location ISO Code,Location,Total Cases,Country,Continent
3/1/2020,ID-JK,DKI Jakarta,39,Indonesia,Asia
3/2/2020,ID-JK,DKI Jakarta,41,Indonesia,Asia
3/2/2020,IDN,Indonesia,2,Indonesia,Asia
3/2/2020,ID-RI,Riau,1,Indonesia,Asia
3/3/2020,ID-JK,DKI Jakarta,43,Indonesia,Asia
\end{filecontents*}
\csvautobooktabular[separator=comma, respect all, filter={\value{csvrow}<3}]{data/covid_dataset_ind_synthetic_head.csv}%
    }
    \vspace{1em}
    \caption{COVID-19 Indonesia}
    \label{tab:ind-samples}
\end{subtable}\\

\begin{subtable}[t]{\textwidth}
    \centering
    \adjustbox{max width=\textwidth}{%
        \begin{filecontents*}{data/covid_dataset_it_synthetic_head.csv}
Date,Country,RegionCode,RegionName,ProvinceCode,ProvinceName,ProvinceAbbreviation,Latitude,Longitude,TotalPositiveCases
2020-02-24T18:00:00,ITA,13,Abruzzo,66,L'Aquila,AQ,42.35122196,13.39843823,0
2020-02-24T18:00:00,ITA,13,Abruzzo,67,Teramo,TE,42.6589177,13.70439971,0
2020-02-24T18:00:00,ITA,13,Abruzzo,68,Pescara,PE,42.46458398,14.21364822,0
2020-02-24T18:00:00,ITA,13,Abruzzo,69,Chieti,CH,42.35103167,14.16754574,0
2020-02-24T18:00:00,ITA,17,Basilicata,76,Potenza,PZ,40.63947052,15.80514834,0
\end{filecontents*}
\csvautobooktabular[separator=comma, respect all, filter={\value{csvrow}<3}]{data/covid_dataset_it_synthetic_head.csv}%
    }
    \vspace{1em}
    \caption{COVID-19 Italy}
    \label{tab:it-samples}
\end{subtable}\\
\begin{subtable}[t]{0.48\textwidth}
    \centering
    \adjustbox{max width=\textwidth}{%
        \begin{filecontents*}{data/covid_dataset_US_synthetic_head.csv}
date,county,state,fips,cases,deaths
2020-01-21,Snohomish,Washington,53061.0,1,0.0
2020-01-22,Snohomish,Washington,53061.0,1,0.0
2020-01-23,Snohomish,Washington,53061.0,1,0.0
2020-01-24,Cook,Illinois,17031.0,1,0.0
2020-01-24,Snohomish,Washington,53061.0,1,0.0
\end{filecontents*}
\csvautobooktabular[separator=comma, respect all, filter={\value{csvrow}<3}]{data/covid_dataset_US_synthetic_head.csv}%
    }
    \vspace{1em}
    \caption{COVID-19 US}
    \label{tab:us-samples}
\end{subtable}
\vspace{-2em}
\end{table*}
\Cref{tab:covid-samples} shows a snapshot of synthetic samples generated from each dataset. A brief inspection of these samples reveals that, despite having several conceptually aligned features, harmonizing them requires substantial pre-processing. Aligning semantically equivalent columns ranges from simple operations, such as standardizing feature names (e.g., converting \texttt{Date} to \texttt{date}), to more complex tasks, such as aggregating multiple location-related features and selecting a consistent level of granularity (e.g., reducing province-level information to the country level). These challenges underscore the need for a reasoning-driven alignment process capable of interpreting semantic meaning rather than relying solely on string matching or manual rules. Moreover, discrepancies in representation formats, such as ISO codes or GPS coordinates, require dataset-specific transformations that must be consistently applied across sites to enable meaningful harmonization.

\noindent\textbf{Models.} Each agent in \NAME{} can either rely on a shared large language model (LLM) or use specialized LLMs tailored to its specific task (e.g., use a coding \ac{LLM} for the code generation of transformations). In our experiments, we evaluate \texttt{GPT-OSS-120b}~\cite{agarwal2025gpt}(\texttt{GPT-120b}) and \texttt{llama-3-3-70b-instruct} (\texttt{Llama-70b})~\cite{dubey2024llama} for dataset analysis, clustering features, and recommending transformations. For code generation, we evaluate the general-purpose \texttt{GPT-120b} and a code generation \ac{LLM} \texttt{deepseek-coder-33b-instruct} (\texttt{DSC-33b})~\cite{guo2024deepseek}.

\noindent\textbf{Results.} 
We evaluate the performance of \NAME{} using multiple combinations of the four datasets. To quantify the alignment efficiency, we use three metrics: (i) the total \# of iterations needed for convergence, (ii) the total \# of transformations recommended during the process, and (iii) the \# of common features -- defined as features sharing the same name and DBpedia URI -- after harmonization. The maximum number of iterations is capped at $T=20$. We first construct all possible combinations, yielding $11$ dataset combinations, and focus our analysis on a subset of these. \Cref{tab:example} shows an example of the harmonized output for the AFG+IDN pair (using \texttt{GPT-120b} for all alignment steps). Compared with the original schema (shown in \Cref{tab:covid-samples}), \NAME{} consistently standardized the feature names to snake case, harmonized ISO codes, and aligned date formats to match the COVID-19 Indonesia convention (see Listing~\ref{lst:code_example} for an example). The framework preserved dataset-specific attributes when appropriate, for instance, retaining \textit{population\_density} from AFG, while removing features that could not be consistently represented across datasets, such as \textit{Location} from IDN, which lacked a viable counterpart in AFG.

\begin{table}[tb]
\caption{Example of AFG and IDN from \textbf{gpt\_all}.}
\label{tab:example}
\centering
\begin{subtable}[t]{.5\textwidth}
    \centering
    \scriptsize
    \adjustbox{max width=\textwidth}{%
       \begin{filecontents*}{data/covid19-dataset_afg_ind_gpt_all.csv}
iso_code,date,country,continent,total_cases,population_density
AFG,2020-02-24,Afghanistan,Asia,5.0,54.422
AFG,2020-02-25,Afghanistan,Asia,5.0,54.422
AFG,2020-02-26,Afghanistan,Asia,5.0,54.422
AFG,2020-02-27,Afghanistan,Asia,5.0,54.422
AFG,2020-02-28,Afghanistan,Asia,5.0,54.422
\end{filecontents*}
\csvautobooktabular[
        separator=comma,
        respect all,
        filter={\value{csvrow}<3}
        ]{data/covid19-dataset_afg_ind_gpt_all.csv}
        }
    \vspace{1.2em}
    \caption{COVID-19 Afghanistan}
    \label{tab:afg-samples-aligned}
\end{subtable}\quad
\begin{subtable}[t]{0.3\textwidth}
    \centering
    \scriptsize
    \adjustbox{max width=\textwidth}{%
        \begin{filecontents*}{data/covid19-indonesia_afg_ind_gpt_all.csv}
iso_code,date,continent,total_cases
IDN,2020-03-01,Asia,39
IDN,2020-03-02,Asia,41
IDN,2020-03-02,Asia,2
IDN,2020-03-02,Asia,1
IDN,2020-03-03,Asia,43
\end{filecontents*}
\csvautobooktabular[separator=comma, respect all, filter={\value{csvrow}<3}]{data/covid19-indonesia_afg_ind_gpt_all.csv}%
    }
    \vspace{1.2em}
    \caption{COVID-19 Indonesia}
    \label{tab:ind-samples-aligned}
\end{subtable}
\vspace{-2em}
\end{table}

\begin{lstlisting}[
    language=Python,
    caption={Example of \texttt{GPT-120b} feature transformation.},
    % recommended by \texttt{GPT-120b}.},
    label={lst:code_example},
    basicstyle=\scriptsize\ttfamily,
    breaklines=true,
    columns=flexible,
    keepspaces=true
]
import pandas as pd

def transform_date(value):
    """
    Convert a date string (e.g., '3/28/2020') into 
    ISO-8601 format 'YYYY-MM-DD'.
    Returns the formatted string, or the original 
    value if parsing fails.
    """
    try:
        # pandas can parse many date formats
        if pd.isna(value):
            return value
        return pd.to_datetime(value).strftime('%Y-%m-%d')
    except Exception:
        return value
\end{lstlisting}

The results of all conducted experiments are summarized in \Cref{tab:model_comparison}, which also shows that the number of aligned features can vary across model configurations for the same dataset combination. For instance, in the IDN-AFG pair, the number of aligned features ranges from $4$ to $5$ depending on the underlying LLM. This variability becomes more obvious in the AFG-IDN-IT triplet, where configurations using \texttt{Llama-70b} result in only 1-2 common features. When moving from pairwise to multi-party harmonization, the number of required transformations increases by 10x, highlighting the growing complexity of aligning heterogeneous schemas. The triplet and all-datasets combinations are particularly demanding, requiring $49$ and $78$ transformations, respectively, showing that multi-party harmonization scales non-linearly in complexity.

\begin{table*}[t]
\centering
\caption{The total number of iterations and transformations applied to harmonize the datasets across various LLMs.}
\label{tab:model_comparison}
\scriptsize
\adjustbox{max width=\textwidth}{
\begin{tabular}{|l|l|c|c|c|c|c|c|c|}
\hline
\textbf{Parties} & \textbf{Experiment} & \textbf{Clust. Feat. $A_C$} & \textbf{Trans. Rec. $A_H$} & \textbf{Code Gen. $M_C$} & \textbf{Dataset Analyzer $A_D$, $A_F$, $A_T$, $A_R$} & \textbf{\# Iter.}  & \textbf{\# Trans.} & \textbf{\# Feat.} \\
\hline
 & \textbf{gpt\_all} & \texttt{GPT-120b} & \texttt{GPT-120b} & \texttt{GPT-120b} & \texttt{GPT-120b} & 2 & 5 & 4 \\
& \textbf{gpt\_all\_ds\_code} & \texttt{GPT-120b} & \texttt{GPT-120b} & \texttt{DSC-33b}  & \texttt{GPT-120b} & 11 & 22 & 4 \\
 \textbf{IDN},\textbf{AFG} & \textbf{gpt\_all\_llama\_da} & \texttt{GPT-120b} &  \texttt{GPT-120b} & \texttt{GPT-120b} & \texttt{Llama-70b}  & 15 & 22 & 5 \\
 & \textbf{gpt\_norm\_code\_llama\_clu\_da} & \texttt{Llama-70b} & \texttt{GPT-120b} & \texttt{GPT-120b} & \texttt{Llama-70b}  & 3 & 8 & 5 \\
& \textbf{llama\_all\_gpt\_code}  & \texttt{Llama-70b} &  \texttt{Llama-70b} & \texttt{GPT-120b} & \texttt{Llama-70b}  & 3 & 14 & 5 \\
\hline
 & \textbf{gpt\_all} & \texttt{GPT-120b} & \texttt{GPT-120b} & \texttt{GPT-120b} & \texttt{GPT-120b} & 4 & 7 & 3 \\
 & \textbf{gpt\_all\_ds\_code} & \texttt{GPT-120b} & \texttt{GPT-120b} & \texttt{DSC-33b}  & \texttt{GPT-120b} & 8 & 24 & 3 \\
\textbf{IDN},\textbf{IT} & \textbf{gpt\_all\_llama\_da} & \texttt{GPT-120b} & \texttt{GPT-120b} & \texttt{GPT-120b} & \texttt{Llama-70b}  & 10 & 27 & 2 \\
 &  \textbf{gpt\_norm\_code\_llama\_clu\_da} & \texttt{Llama-70b} & \texttt{GPT-120b} & \texttt{GPT-120b} & \texttt{Llama-70b}  & 2 & 6 & 3 \\
  & \textbf{llama\_all\_gpt\_code}& \texttt{Llama-70b} & \texttt{Llama-70b} & \texttt{GPT-120b} & \texttt{Llama-70b}  & 19 & 38 & 4 \\
\hline
 & \textbf{gpt\_all} & \texttt{GPT-120b} & \texttt{GPT-120b} & \texttt{GPT-120b} & \texttt{GPT-120b} & 12 & 32 & 3 \\
 & \textbf{gpt\_all\_ds\_code} & \texttt{GPT-120b} & \texttt{GPT-120b} & \texttt{DSC-33b}  & \texttt{GPT-120b} & 12 & 61 & 2 \\
\textbf{AFG},\textbf{IDN},\textbf{IT}& \textbf{gpt\_all\_llama\_da} & \texttt{GPT-120b} & \texttt{GPT-120b} & \texttt{GPT-120b} & \texttt{Llama-70b}  & 4 & 15 & 3 \\
 &  \textbf{gpt\_norm\_code\_llama\_clu\_da} & \texttt{Llama-70b} & \texttt{GPT-120b} & \texttt{GPT-120b} & \texttt{Llama-70b}  & 18 & 35 & 4 \\
 & \textbf{llama\_all\_gpt\_code}& \texttt{Llama-70b} & \texttt{Llama-70b} & \texttt{GPT-120b} & \texttt{Llama-70b}  & 20 & 49 & 1 \\
\hline
 & \textbf{gpt\_all} & \texttt{GPT-120b} & \texttt{GPT-120b} & \texttt{GPT-120b} & \texttt{GPT-120b}  & 20 & 78 & 1 \\
 & \textbf{gpt\_all\_ds\_code} & \texttt{GPT-120b} & \texttt{GPT-120b} & \texttt{DSC-33b}  & \texttt{GPT-120b}  & 5 & 31 & 2 \\
\textbf{ALL} & \textbf{gpt\_all\_llama\_da} & \texttt{GPT-120b} & \texttt{GPT-120b} & \texttt{GPT-120b} & \texttt{Llama-70b}  & 3 & 18 & 2 \\
 &  \textbf{gpt\_norm\_code\_llama\_clu\_da} & \texttt{Llama-70b} & \texttt{GPT-120b} & \texttt{GPT-120b} & \texttt{Llama-70b}  & 2 & 15 & 1 \\
 & \textbf{llama\_all\_gpt\_code}& \texttt{Llama-70b} & \texttt{Llama-70b} & \texttt{GPT-120b} & \texttt{Llama-70b}  & 4 & 22 & 3 \\
\hline
\end{tabular}
}
\end{table*}

\begin{figure}[tb]
    \vspace{-10pt}
    \centering
    \begin{subfigure}[c]{0.22\textwidth}{\includegraphics[width=\textwidth]{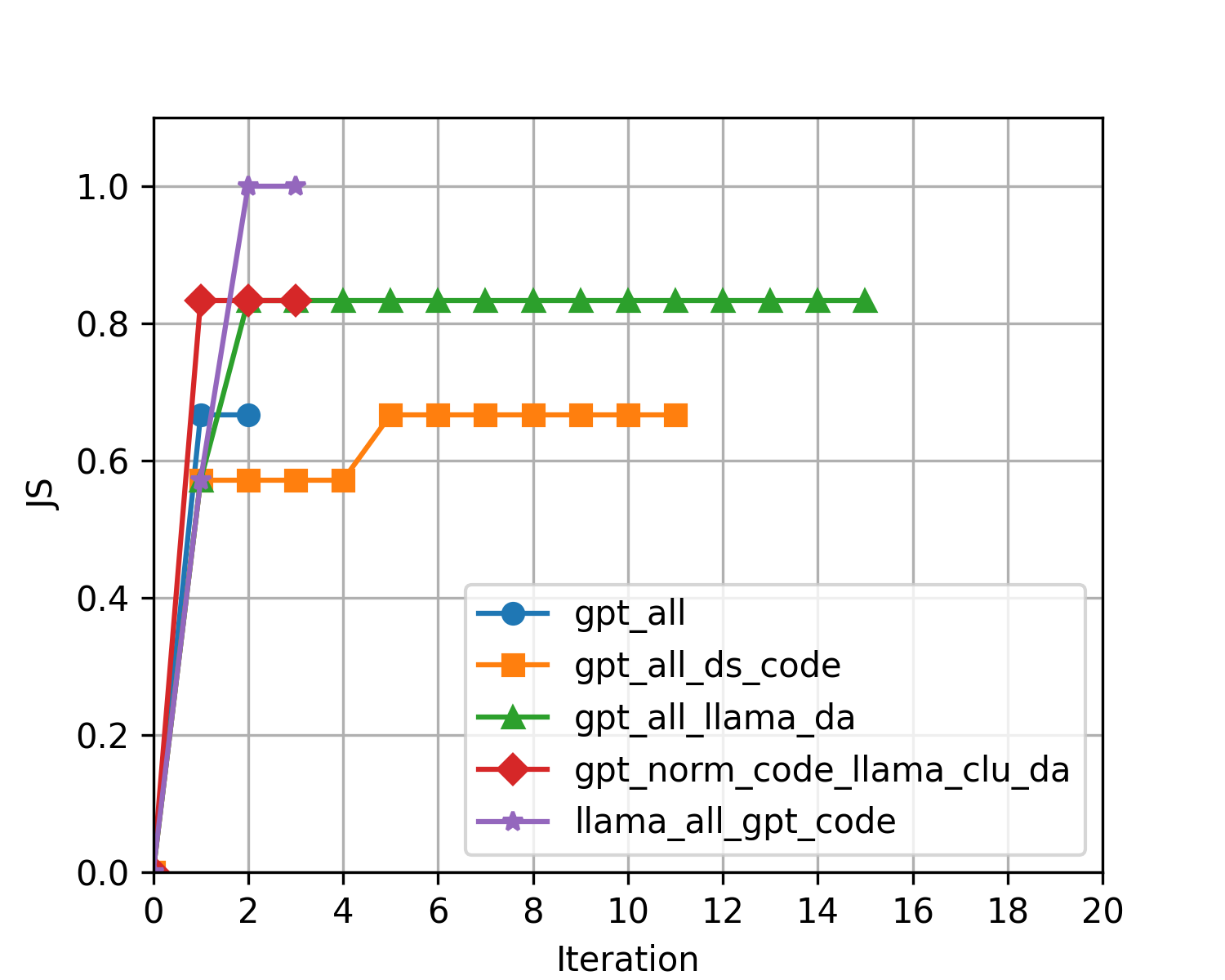}}
    \caption{IDN + AFG}
    \end{subfigure}\qquad
    \begin{subfigure}[c]{0.22\textwidth}{\includegraphics[width=\textwidth]{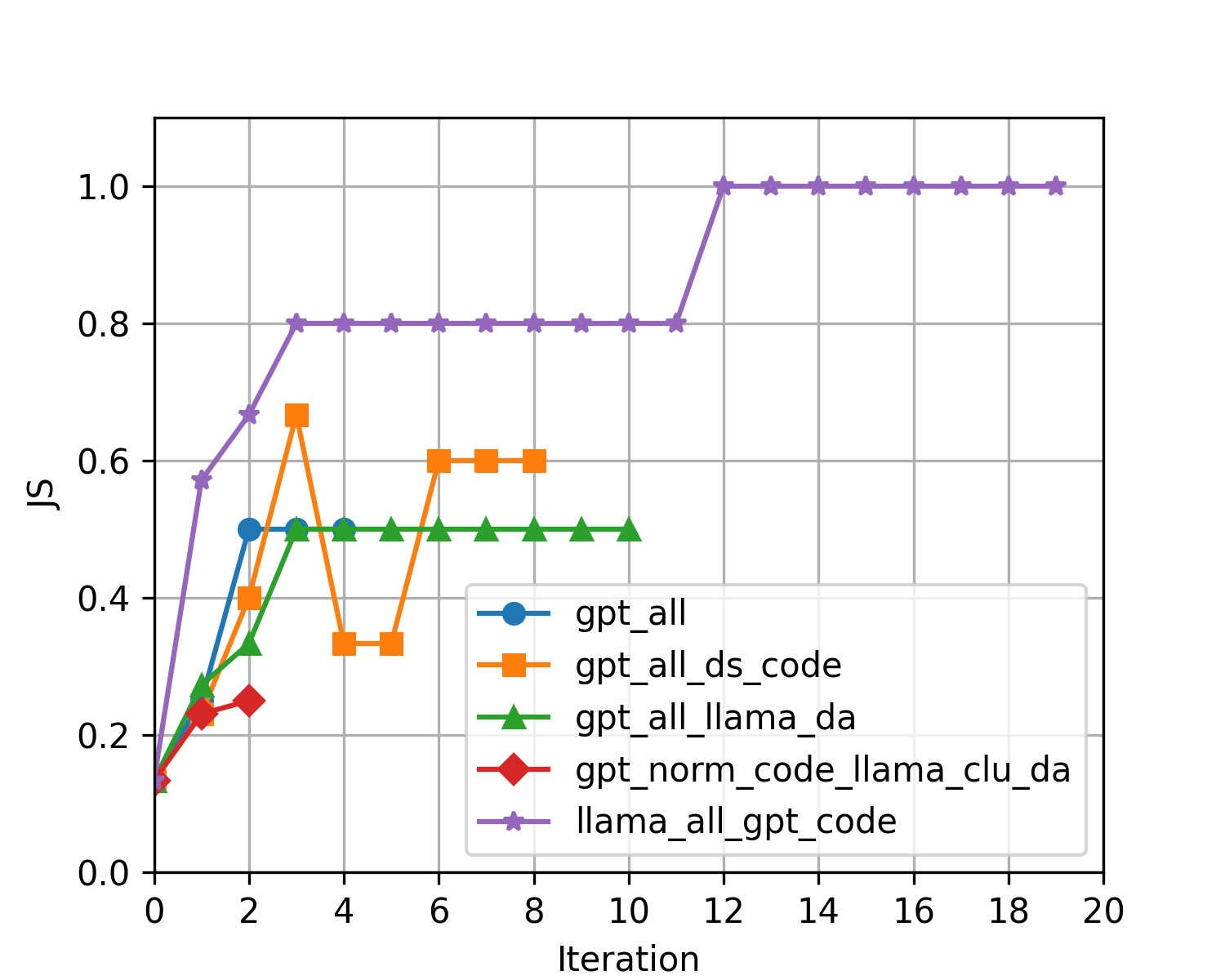}}
    \caption{IDN + IT}
    \end{subfigure}\qquad
    \begin{subfigure}[c]{0.22\textwidth}{\includegraphics[width=\textwidth]{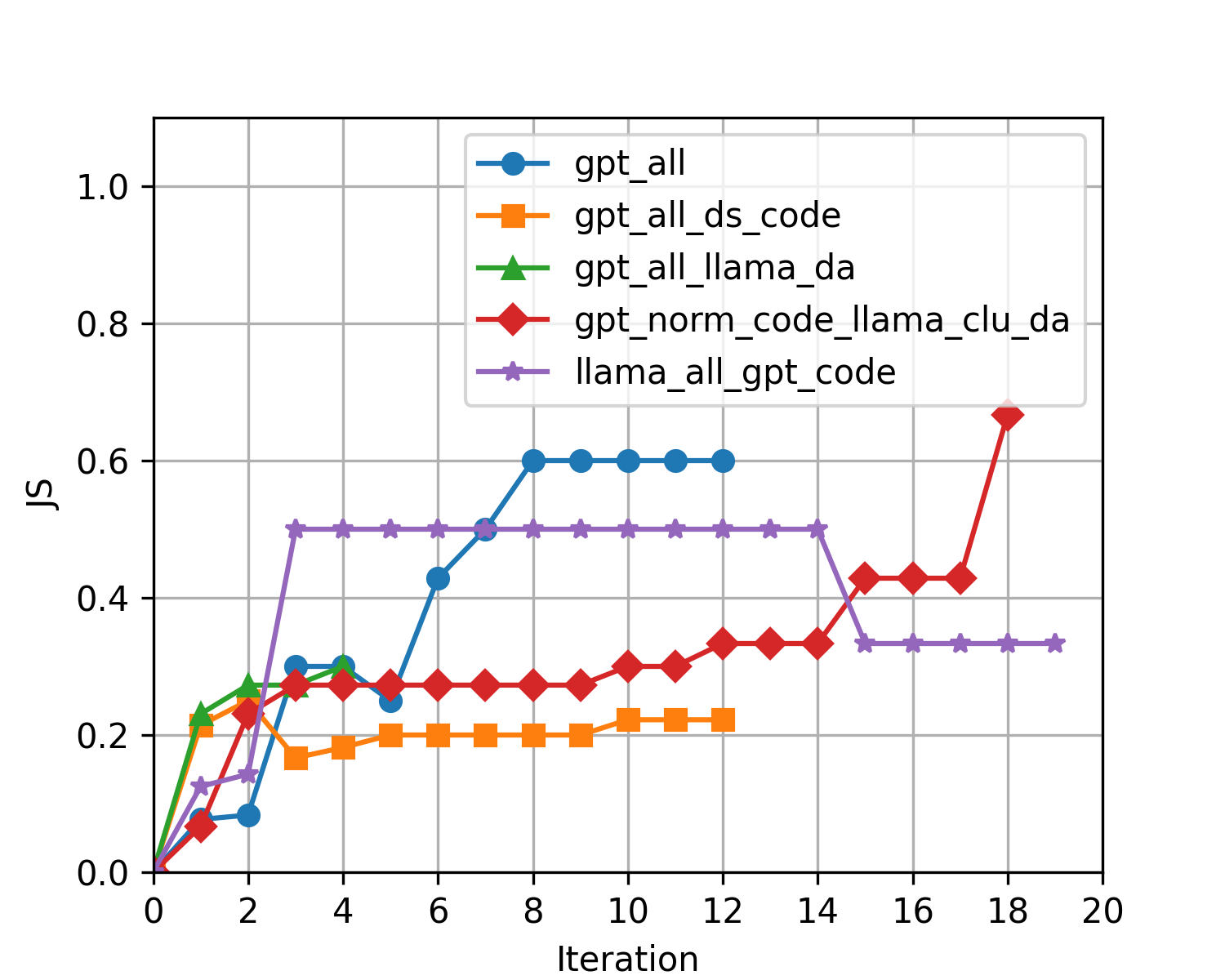}}
    \caption{AFG + IT + IDN }
    \end{subfigure}\qquad
    \begin{subfigure}[c]{0.22\textwidth}{\includegraphics[width=\textwidth]{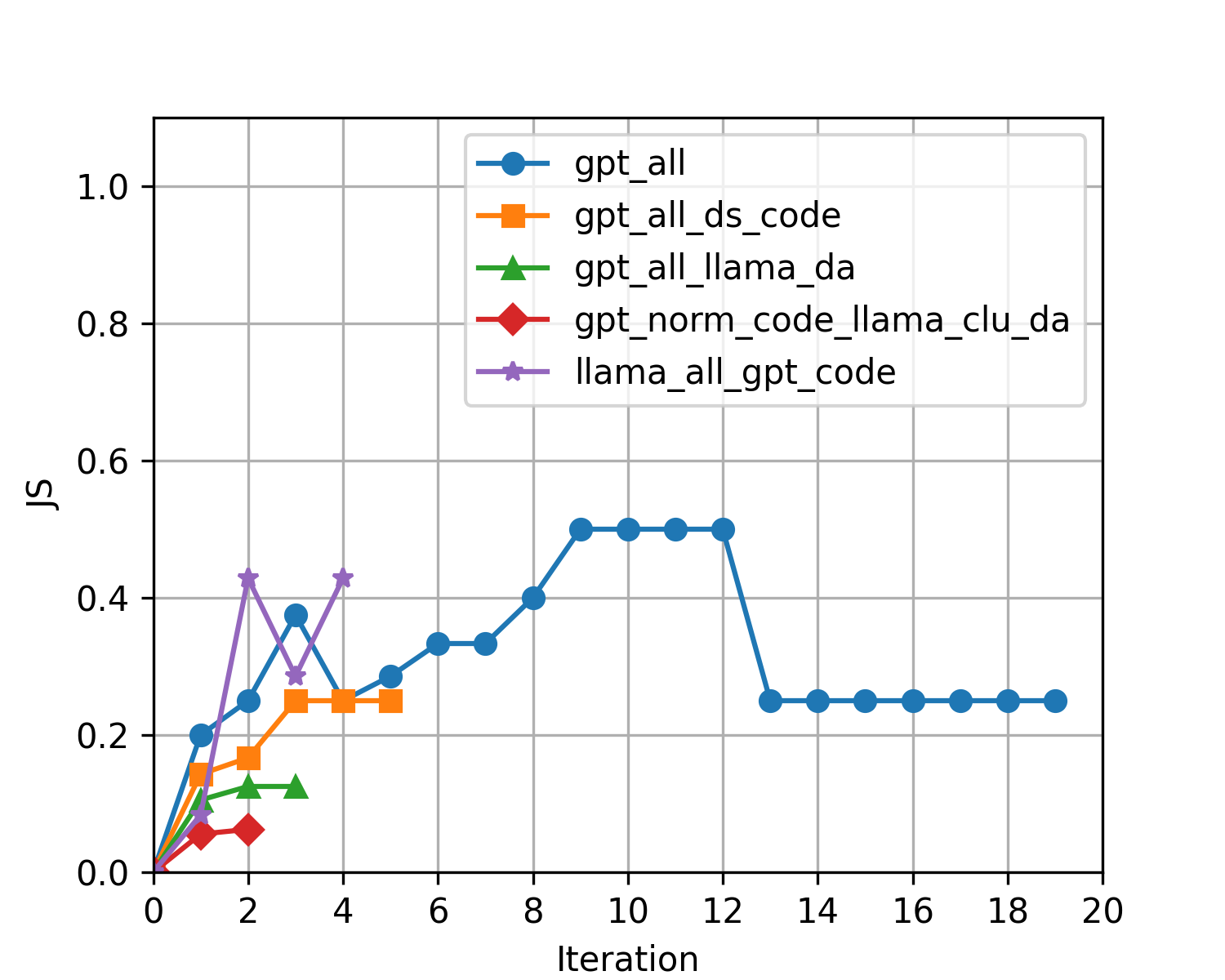}}
    \caption{IT + IDN + AFG + US}
    \end{subfigure}\qquad
    \caption{Jaccard Similarity wrt iterations \# for different LLMs.}
    \vspace{-10pt}
    \label{fig:results-similarity}
\end{figure}

To assess the degree of alignment, we compute the \ac{JS}~\cite{jaccard1901} between feature names at each iteration. This metric quantifies the proportion of shared features (those with matching names and formats) relative to the total \# of unique features across the two datasets. Figure~\ref{fig:results-similarity} shows the \ac{JS} increasing with each iteration for all models and dataset combinations. Across all subplots, most configurations reach their peak \ac{JS} within 3–6 iterations, indicating quick stabilization. Most improvements occur in the first 2–4 iterations, after which curves flatten, highlighting the importance of early iterations. \textbf{gpt\_all} and \textbf{gpt\_all\_ds\_code} typically stabilize around JS~$\in (0.4, 0.6)$, which is significantly lower than LLaMA-based results. We noticed that \texttt{Llama-70b} tends to introduce new features, i.e., \textit{Disease} and fill it with \textit{COVID-19}, which explains the higher $JS$ scores for LLaMA combinations.

As part of its design, \NAME{} is instructed to preserve dataset-specific features when they can not be generated, inferred, or reliably aligned across sites. This behavior explains the lower scores observed in some model combinations, where agents were unable to cluster or map certain features due to their uniqueness. Conversely, for the AFG+IDN pair, in the \textbf{afg\_ind\_llama\_all\_gpt\_code} configuration, where \texttt{Llama-70b} was used for all stages except code generation, the framework proposed a transformation that removed the \textit{population\_density} feature from AFG due to its uniqueness in the dataset (see \Cref{tab:covid-samples} for a snapshot of the dataset).

%% file: sections/discussion.tex
\section{Discussion}\label{sec:discussion}

\noindent\textbf{Impact of \ac{LLM}s.} Our evaluation revealed distinct performance characteristics across the tested \acp{LLM}. For example, GPT-120b demonstrated superior instruction adherence while maintaining creativity, often recognizing when further iterations were unnecessary, even in independent optimization cycles. The experiment referenced as \textbf{gpt\_all} required fewer iterations and transformations, favoring incremental updates over introducing new features. In contrast, LLaMA exhibited higher creativity but occasionally misapplied normalization steps or introduced unnecessary changes (e.g., renaming features or altering string cases). While LLaMA performed well in data analysis tasks, it struggled in determining when the harmonization process should terminate. DeepSeek consistently followed instructions and produced correct code. However, it more frequently generated incomplete responses requiring manual completion (e.g., placeholder lists with comments). Both GPT-120b and DeepSeek occasionally violated coding guidelines by prematurely replacing values with None before attempting transformations. Finally, combining LLaMA with DeepSeek proved ineffective, causing persistent errors and non-convergence within $T=20$ iterations.

\noindent\textbf{Privacy Analysis of \NAME{}.} Under the threat model (Section~\ref{sec:system_threat_model}), we assume an honest-but-curious setting in which the server and participating researchers follow the prescribed protocol, and no parties collude to extract additional information about any specific participant. Within this context, we can analyze the steps executed by \NAME{} and conclude that no unintended information is leaked. Each participant shares only high‑level ``structural metadata'' (data types, semantic feature descriptions) and a small set of synthetic samples. Researchers fully control synthetic sample generation, and \NAME{} requires only syntactically valid, not statistically faithful, examples. This allows correlation‑free generation that lowers re‑identification risk~\cite{ganev2025inadequacy}. Each researcher receives only the transformation instructions for their own dataset, preventing visibility into others’ data characteristics. While the transformation instructions may reveal the overall level of granularity or formatting chosen during harmonization, this information reflects an aggregate view across datasets rather than any single researcher's contribution. Because the transformations are produced as a collective abstraction (coordinated by the server), participants cannot reliably attribute specific schema elements or granularity decisions to individual researchers. Overall, by restricting communication to structural metadata and locally generated, low-utility synthetic samples, and by limiting the visibility of transformation recommendations, \NAME{} effectively minimizes cross-site disclosure risks while enabling automated, privacy-preserving harmonization.

\noindent\textbf{Limitations.} While \NAME{} is designed to be adaptable to a wide range of domains, several limitations remain. Its performance depends on the reasoning consistency of \ac{LLM}, which may vary across models and tasks. Although the framework encourages preservation of dataset-specific features, LLM-driven harmonization may occasionally over/under-align features. In clinical settings, incorporating a human-in-the-loop to review and validate the harmonization recommendations would strengthen semantic accuracy and reliability. Additionally, \NAME{} typically converged in our experiments, but lacks theoretical guarantees on iteration count. Finally, the multi-agent workflow can be computationally intensive and may not be feasible for resource-constrained institutions. 

%% file: sections/related_work.tex
\section{Related Work}
\label{sec:related-work}
We review two primary lines of related research: (i) data harmonization and (ii)federated and distributed health analytics.

\noindent\textbf{Data Harmonization.} 
Data harmonization has been extensively studied across a range of scientific domains, often in field-specific contexts such as genomics, epidemiology, and multi-omics research~\cite{kumar2021data,zhu2019challenges,chen2018ontology,chen2023multiomics}. Broadly, harmonization approaches can be classified into retrospective and prospective strategies~\cite{cheng2024general}. Retrospective harmonization aligns datasets that have already been collected, as in the case of harmonizing existing COVID-19 datasets~\cite{hurtt2011harmonization}. In contrast, prospective harmonization requires collaborators to agree on standardized measures prior to data collection. While powerful, this approach can be challenging to implement because research groups often operate under different scientific aims, or healthcare systems~\cite {uphoff2003harmonisation}. In practice, harmonization efforts often fall along a continuum between these two extremes~\cite{torres2022harmonization}. Harmonization techniques can also be classified by the mechanism used to align heterogeneous data. One strategy is merging, in which a unified taxonomy or ontology is constructed that encompasses all local taxonomies or ontologies. Alternatively, mapping-based methods define alignment rules between ontologies, enabling interoperability without requiring complete unification.
These approaches offer different trade-offs in scalability, flexibility, and consistency, depending on the intended analytical use case~\cite{cheng2024general}. 

\noindent\textbf{Federated and Distributed Health Analytics.} 
Several prior works have focused on enabling multi-institutional health analytics and collaborative model development without centralizing sensitive data~\cite{gaye2014datashield,deist2020personalhealthtrain}. Gaye et al.~\cite{gaye2014datashield} develop DataShield, a framework that enables the co-analysis of individual-level data from several studies without transferring the data across institutions. Similarly, the \ac{PHT}~\cite{deist2020personalhealthtrain} provides a distributed infrastructure that brings computation to the data rather than the reverse. Evaluations on lung cancer datasets (tumor staging and post-treatment survival information) have shown the effectiveness of \ac{PHT}. More recent works in \ac{FL} have explored how to improve model performance under heterogeneous client data distributions. Personalized FL approaches aim to tailor global models to local client needs~\cite{tan2022towards}. Per-FedAvg~\cite{fallah2020personalized} uses a meta-learning strategy to learn an initial shared model, which clients can rapidly adapt to their local distribution with only a few gradient updates. Li et al.~\cite{li2021sample} propose a privacy-preserving method for selecting informative training samples, enabling clients to build more effective models while reducing computation.

%% file: sections/conclusions.tex
\section{Conclusion}\label{sec:conclusion}
In this work, we have proposed \NAME{}, a novel multi-agent framework for harmonizing datasets across sites in a privacy-preserving way. Evaluation on four real-world COVID-19 datasets showed that \NAME{}, particularly when leveraging GPT-OSS as the underlying large language model, effectively harmonized datasets in just a few iterations with a minimal number of required transformations. By enabling automatic alignment of heterogeneous datasets while preserving data privacy, \NAME{} will facilitate high-quality collaborative research across institutions. In future work, we will explore the application of \NAME{} to additional datasets and domains, as well as the evaluation of alternative large language models to further enhance its scalability and applicability.